\documentclass[
]{ceurart}

\usepackage{graphicx}
\usepackage{dirtytalk}
\usepackage{selinput}
\usepackage{selinput}
\usepackage[spanish]{babel}

\begin{document}

\copyrightyear{2020}
\copyrightclause{Copyright for this paper by its authors.
  Use permitted under Creative Commons License Attribution 4.0
  International (CC BY 4.0).}

\conference{Fire 20: Forum for Information Retrieval Evaluation, December 16–20, 2020,  Hyderabad, India}

\title{Overview of the Shared Task on Fake News Detection in Urdu at FIRE 2020}

\author[1]{Maaz  Amjad}[%
%orcid=0000-0002-0877-7063,
email=maazamjad@phystech.edu,
url=https://nlp.cic.ipn.mx/maazamjad/,
]
\address[1]{Center for Computing Research (CIC), Instituto Politécnico Nacional (IPN), Mexico }

\author[1]{Grigori Sidorov}[%
%orcid=0000-0001-7116-9338,
email=sidorov@cic.ipn.mx,
]

\author[2]{Alisa Zhila}[%
%orcid=0000-0002-9421-8566,
email=alisa.zhila@gmail.com,
%url=http://conceptbase.sourceforge.net/mjf/,
]
\address[2]{Independent Researcher, United States}

\author[1]{Alexander Gelbukh}[%
%orcid=0000-0002-9421-8566,
email=gelbukh@gelbukh.com,
%url=http://conceptbase.sourceforge.net/mjf/,
]

\author[3]{Paolo Rosso}[%
%orcid=0000-0002-9421-8566,
email=prosso@dsic.upv.es,
%url=http://conceptbase.sourceforge.net/mjf/,
]
\address[3]{Universitat Politècnica de València, Spain}

\begin{abstract}
 This overview paper describes the first shared task on fake news detection in Urdu language. The task was posed as a binary classification task, in which the goal is to differentiate between real and fake news. We provided a dataset divided into 900 annotated news articles for training and 400 news articles for testing. The dataset contained news in five domains: (i) Health, (ii) Sports, (iii) Showbiz, (iv) Technology, and (v) Business. 42 teams from 6 different countries (India, China, Egypt, Germany, Pakistan, and the UK) registered for the task. 9 teams submitted their experimental results. The participants used various machine learning methods ranging from feature-based traditional machine learning to neural networks techniques. The best performing system achieved an F-score value of 0.90, showing that the BERT-based approach outperforms other machine learning techniques.
\end{abstract}


\maketitle

\section{Introduction}

Fake news dissemination has been an important issue starting in the 15th\footnote{https://www.blackbird.ai/history-of-misinformation/} century. While meant to be objective, not all news articles follow the rigor of conveying fair facts chasing the fast-paced readers' attention by screaming headlines and sensational content. The spread of fake news brought many technical and social challenges. For example, it was a dispersion of fake news, which ignited the origin of antisemitism\footnote{ https://libguides.ncl.ac.uk/fakenews/history} in 1475, when a Franciscan preacher on the occasion of Easter Sunday claimed that Jewish community killed a toddler. Moreover, to celebrate the child’s pass-over, some Jewish drained the child's blood and drank. The fake news spread fast and as a revenge, Trent's whole Jewish community was arrested, tortured, and fifteen Jewish were found guilty and burned at the stake. This story inspired surrounding local communities to commit similar atrocities. Thus, propagation of fake news brought terrifying results and inflamed social conflict. 

Automatic fake news identification is difficult, because we deal with very high level semantic phenomenon and at the first glance fake news look like real news. There are several types of news that are considered fake news. The researchers \cite{2} classified fake news into six types: (i) fabrication, (ii) news satire, (iii) manipulation (e.g., editing pictures), (iv) advertising (e.g., ads are depicted as professional journalism), (v) propaganda, and (vi) news parody. Fabricated news can be defined as a deliberately fabricated news article manipulated to deliver a particular narrative, such as to create confusion, to be prominent in news headlines, or to make money.

The term  ``\verb|fake news|'' is not a simple concept. Publishers have been spreading false and misleading information even before the availability of the Internet. Different studies proposed various definitions of fake news \cite{1, 2, 3}.  For example, a recent study \cite{3} defined fake news as a factually incorrect news article, which intentionally misleads a reader to believe that the conveyed information is true. There is a related term \textit{clickbait}, which is defined as a snippet of a news article that is used to attract the reader's attention, and upon clicking, it redirects the reader to a different page. Clickbaits are used to generate revenues by online advertisements. Notably, the term \textit{misinformation} – spreading untruths – emerged in the late 16th century\footnote{https://www.theguardian.com/books/2019/nov/22/factitious-taradiddle-dictionary-real-history-fake-news }. Further, the more specific term \textit{disinformation} came from Russian word\footnote{https://www.theguardian.com/books/2019/jun/19/dominic-raab-disinformation} “dezinformacija”. Disinformation is used when there is the intent to harm. The Guardian showed that this term was used during the cold war by all participants, being its meaning sowing falsehoods to confuse enemies.

Automatic  detection of fake news is crucial to prevent the devastating and havoc impact that the fake news phenomenon causes worldwide. Throughout history, humans have always been prejudiced and intolerant to different views. For example, in 1620, Francis Bacon\footnote{ https://www.theguardian.com/technology/2016/nov/29/fake-news-echo-chamber-ethics-infosphere-internet-digital } emphasized the consequences of inaccurate language in his book \textit{Novum Organum}, “The ill and unfit choice of words wonderfully obstructs the understanding.” To this point, fake news detection is a means to eliminate the vast and disastrous effects produced by misinformation. Further, the ever increasing pace and scale of fake news propagation can be mitigated only by automating the solutions and increasing their effectiveness.  

Fortunately, fake news detection attracted many researchers, in particular after the US 2016 presidential election. The importance of the task visibly increased for the Natural language processing (NLP) community as well as the demand for the solutions from the industry. Given the topicality of fake news detection and the urgency of coming up with an effective solution, this competition aims to gain the attention of a larger research community and to incentivize development of different  solutions to combat  the propagation of fake news on the Web.

\section{Importance of Fake News Detection in Urdu}

A large number of existing studies in the  literature examined automatic fake news detection in multiple languages,  such as English, Spanish, German, Chinese, and Arabic. However, a very limited work was done in Urdu language to automatically identify fake news content on the web. Urdu is a widely spoken language having more than 100 million native speakers  worldwide\footnote{ http://www.bbc.co.uk/voices/multilingual/urdu.shtml}. However, according to the best of our knowledge,  there are no automatic web sources to verify the authenticity of news articles in Urdu language. This situation requires the attention of researchers working in NLP to develop tools and solutions in verifying the authenticity of news articles written in Urdu language. Note that Urdu is a low resource language, i.e., it does not have many NLP tools and corpora data.

Urdu is spoken mainly in Pakistan. The fake news phenomenon had bad effects in Pakistan's social, politics and economical situation. For example, a Pakistani TV anchor Dr. Shahid Masood\footnote{ https://www.globalvillagespace.com/dr-shahid-masoods-claims-about-zainabs-murderer-prove-false/ }, was sent to jail and barred from hosting the TV show due to deceptive claims and spreading fake news on the rape case of a teenage girl during a television show. 
Similarly, according to the Washington Post\footnote{https://www.washingtonpost.com/world/asia\_pacific/as-mob-lynchings-fueled-by-whatsapp-sweep-india-authorities-struggle-to-combat-fake-news/2018/07/02/683a1578-7bba-11e8-ac4e-421ef7165923\_story.html}, fake news about child trafficking led to many deaths of innocent people in India. 

In addition to this,  BBC reported that some Indian sites claimed a civil war\footnote{ https://www.bbc.com/news/world-asia-54649302 } had broken out in one of the cities of Pakistan. The report mentioned that some Indian websites described the situation in Pakistan as dangerous and the civil war resulted in the deaths of many city police officers. Moreover, the websites claimed that tanks had been seen on the streets, which eventually proved to be fake news. Therefore, this urge to conduct studies to combat the dissemination of  fake content in Urdu language.

\section{Literature Review and Task Overview}

\subsection{Related Work}
A number of approaches for automatic fake news detection were proposed. These approaches are based on statistical text analysis to tackle fake news detection. Previous studies used various datasets, which comprised mainstream media news articles and news published on social media. Notably, the majority of work focused on English \cite{4, 6}, with some efforts in other languages such as Spanish \cite{4, 7}, German \cite{8}, Arabic \cite{5,9},  Persian \cite{10}, Indonasian \cite{11}, Bangla \cite{12},  Portuguese \cite{13}, Dutch \cite{14}, Italian \cite{15}, and Hindi \cite{16}.  

Correspondingly, challenges\footnote{ http://www.fakenewschallenge.org/ } and shared tasks on automatic fake news detection were proposed, such as SemEval 2017 task 8 RumourEval for English \cite{17},  SemEval-2019 task 7 RumourEval for English \cite{18}, and PAN 2020\cite{4}, the latter focused on the author profiling of Fake News Spreaders on Twitter. Moreover, previous studies \cite{20, 21} have shown that emotional information can be helpful to identify fake news on social media effectively. To the best of our knowledge, this is the first shared task on fake news detection for the Urdu language. This task incentivizes the development of fake news detection in Urdu as well as  provides an opportunity to compare the system performances with the recent shared tasks in other languages.

Several studies \cite{6, 7}  reported that linguistic features can be helpful to identify fake news. For example, a recent study \cite{6} demonstrated that  linguistic features proved to be helpful to capture the differences of writing styles between fake and real news. Another study \cite{7} exploited some linguistic features and showed that these features provided cues for differentiating between  fake and real news.

\subsection{Task Description}
The task of fake news detection is to solve a binary classification problem, in which the input is a news article   and the output is the assigned label (real or fake). Built around the idea reported in recent studies that the textual content can be helpful to identify fake news \cite{6, 7}, this shared task is aimed to explore the efficiency of  models to detect fake news, in particular, for the news articles written in Urdu language.

The task was made publically available\footnote{https://www.urdufake2020.cicling.org/home } on June 30, 2020, and the same day the training dataset was released as well. The training dataset was splitted into two subfolders, (i) real news subfolder, and (ii) fake news subfolder. The real news subfolder contained 500 real news articles, and the fake news subfolder contained 400 fake news articles.  We released the test dataset on August 31, 2020. Like for the training dataset, the testing dataset was also splitted into two subfolders, (i) real news, and (ii) fake news. The real news subfolder contained 250 real news articles, and the fake news subfolder contained 150 fake news articles. The participating teams submitted their system until September 10, 2020. Each team could submit up to 3 different runs.

\section{Dataset Collection and Annotation}

This section provides an overview of the dataset developed for the shared task. A smaller version of the dataset, named “Bend The Truth”, along with the detailed information about the collection and annotation description was presented in the recent study \cite{3}. For this task, we performed additional data collection and annotation following the exactly same procedure. As a result, we obtained the final dataset of the annotated fake and real news in Urdu that was 1.5 times the size of the original "Bend-The-Truth" dataset. It is publicly available for academic research\footnote{ https://github.com/UrduFake/urdufake2020eval.git}.

The fake news articles were intentionally written by hired professional journalists under specific instructions.  The domains of the news present in our dataset are: (i) Business, (ii) Health, (iii) Showbiz (entertainment), (iv) Sports, and (v) Technology.  They are similar to the dataset \cite{6} used to identify fake news in English language. Nonetheless,  only one  domain of news, namely, related to education, is not available in our dataset because  it was difficult to find enough verifiable news in Urdu related to the education domain. 

For the training and development purposes, we offered 900 news articles from the previously published “Bend The Truth” dataset. The 400 news articles from the previously unseen and unpublished newly collected part were held out as  a test dataset.

\subsection{Dataset Annotation Procedure}

For the dataset annotation, rigorous guidelines and annotation procedures were defined. The news articles were annotated into two types of news: (i) real news article, and (ii) fake news article. The dataset can be used as a corpus for supervised machine learning. It is possible to use the knowledge of the dataset annotation procedure for applying to the underlying characteristics of fake news in addition to linguistics features, but we do not recommend it, because in real life there is no such information. We followed different strategies for real and fake news annotation. 

\subsubsection{Real News Annotation}
For the annotation of real news articles, initially numerous news articles from different mainstream were collected. Table \ref{tab:reliablesites} shows news agencies used to collect news articles for annotation.  To annotate a news article as a real news, the major points in the real news data collection and annotation were:

\begin{enumerate}
\item The data collection and annotation procedure were performed manually.
\item The news article was labeled as real news if the news meets the following criteria:
\begin{itemize}

\item A reliable newspaper or a prominent news agency published that news article.
\item Other authentic and credible newspaper agencies published the same news article and the veracity of the news article can be easily verified using information such as place of the event, image, date, etc.
We also performed manual source verification from where the news are originated. We further compared and cross checked different sources (mainstream news agencies) to verify the information present in the news article.
\item We also confirmed that a news article has a correlation between its title and its content. We read the complete news articles to find out the correlation between the title and the content. 
\end{itemize}
\end{enumerate}

If a news article does not follow one of these criteria, we simply discard that news article.

Note that the length of all the news articles is heterogeneous. The reason is that each news agency has a different style of news articles. For example, BBC Urdu contained on average more than 1,500 words in a news article. Thus, we selected real news articles carefully following the described procedure.  

\subsubsection{Professional Crowdsourcing of Fake News}
    
To obtain fake news in this dataset, we used professional journalist services from various news agencies in Pakistan: Express news, Dawn news, etc., who were asked to write fake news stories that correspond to the original real news articles. This is a peculiar attribute of this dataset, because it ensured that the fake news articles realistically imitated real life approach to fake news creation. In real life, it is journalists who are responsible for writing fake news articles. Obviously, not all journalists do it. Still, people of this profession have a better understanding of how to write an article (real or fake) and make it interesting to hook and, in case of fake news, trick the reader.

The reasons to use professional “crowdsourcing” to collect fake news are the following:

\begin{enumerate}
\item Finding and verifying the falsehood of fake news in the same domain as the available real news articles is a challenging task that requires a huge amount of time and resources unavailable to a small group of  organizers. Thus, manual analysis of hundreds of thousands of news articles for verification through web scraping approach was unfeasible.
\item Unlike the case of the English language, most of the news verification in Urdu language is done manually due to the absence of web services that offer  news validation.

\end{enumerate}
We should mention that the news articles style and language characteristics vary depending on the news domain. Our dataset contains news in five major domains: sports, business, education, technology, and showbiz. Thus, we assigned news articles according to the journalists expertise in the corresponding domain. Moreover, all the journalists were given instructions to minimize the possibilities of introducing defined patterns that can provide undesirable clues in the classification task. Also, some technical guidelines, such as the requirement that the lengths of fake news should be in the range of those of the original news, were provided. Finally, all  fake news articles were prepared using  journalists’ expertise.

\begin{table}[htb!]
	\caption{Legitimate websites}
	\label{tab:reliablesites}
	 \centering
		\begin{tabular}{ lll }
			\hline\noalign{\smallskip}			
			\multicolumn{1}{c}{\textbf{Name}}& \multicolumn{1}{c}{\textbf{URL}} & \textbf{Origin} \\
			\noalign{\smallskip}\hline\noalign{\smallskip}		
            \textbf{BBC News}& \url{www.bbc.com/urdu} & England\\
            \textbf{CNN Urdu}& \url{cnnurdu.us}& USA\\
           \textbf{Dawn news} & \url{www.dawnnews.tv}	& Pakistan\\
            \textbf{Daily Pakistan} & \url{dailypakistan.com.pk}& Pakistan\\
            \textbf{Eteemad News} & \url{www.etemaaddaily.com} & India\\
            \textbf{Express-News} & \url{www.express.pk} &	Pakistan\\
            \textbf{Hamariweb}&\url{hamariweb.com}&     Pakistan\\
            \textbf{Jung News} & \url{jang.com.pk}&	Pakistan\\
            \textbf{Mashriq News}	& \url{www.mashriqtv.pk} &	Pakistan\\
            \textbf{Nawaiwaqt News} & \url{www.nawaiwaqt.com.pk} & Pakistan\\
            \textbf{Roznama Dunya} &\url{dunya.com.pk}	&	Pakistan \\
            \textbf{The daily siasat} &\url{urdu.siasat.com}	& India \\
            \textbf{Urdu news room} & \url{www.urdunewsroom.com} &	USA\\
            \textbf{Urdupoint} & \url{www.urdupoint.com} & Pakistan\\
            \textbf{Voice of America } & \url{www.urduvoa.com}	& USA\\
            \textbf{Waqt news} &\url{waqtnews.tv} &	Pakistan \\
			\noalign{\smallskip}\hline
		\end{tabular}
\end{table}

\subsection{Training and Testing Datasets}		

\subsubsection{Training and Validation Dataset}

The training set was made available to the participants to develop their approaches to identify fake news. It contained 900 news articles, annotated in a binary manner as real or fake. 500 news articles were annotated as real, and 400 articles were annotated as fake. The real news part of the dataset was retrieved from January 2018 to December 2018, which is different from the test set. 

All  five topic domains, i.e., (i) Business, (ii) Health, (iii) Showbiz (entertainment), (iv) Sports, and (v) Technology  were present in the training dataset. That is, we did not hold out any domain from the training set to make it “unseen” for the participants.  

The use of the training set for validation, development, and parameter tuning purposes was at the participants’ discretion.

\subsubsection{Test dataset}
The test dataset was used to evaluate the performance of the submitted classifiers. It was provided to all the participants without the ground truth labels. The truth labels were only used by the organizers to evaluate and compare the performance of participants’ approaches.

To create the testing dataset, news articles were retrieved from January 2019 to June 2020. It also has all five types of news as the training set. The test dataset is composed of 400 news articles. The ground truth distribution among these 400 news articles was 250 real news articles and 150 fake news articles. We emphasize again that this information, along with  labels, was not made available to the participants.

\subsection{Dataset Statistics}

To prepare the data for the experiments, the corpus was split into train and test sets. In the first stage of the shared task, the training dataset was released which contained 900 news samples (500 real and 400 fake). In the second stage, we released the test dataset which contained 400 news articles (250 real and 150 fake).  Table \ref{tab:Corpus_distribution} describes the corpus distribution of the news articles by topics for the training and testing sets.

\begin{table}[htb!]
	\caption{Domain Distribution in Train and Test subsets}
	\label{tab:Corpus_distribution}
	  \centering
		\begin{tabular}{ lcccc }
			\hline\noalign{\smallskip}			
			\multirow{2}{*}{\textbf{Domain}}& \multicolumn{2}{c}{\textbf{Train}} & \multicolumn{2}{c}{\textbf{Test}}\\
			& \textbf{real} & \textbf{fake}& \textbf{real} & \textbf{fake} \\
			\noalign{\smallskip}\hline\noalign{\smallskip}
			\textbf{Business} & 100& 50 & 50	& 30\\
			\textbf{Health} & 100& 100 & 50	& 30\\		
			\textbf{Showbiz}& 100& 100 & 50	& 30\\
			\textbf{Sports} & 100& 50 & 50	& 30\\
			\textbf{Technology} & 100& 100 & 50	& 30\\
			\noalign{\smallskip}\hline\noalign{\smallskip}
			\textbf{Totals} &\textbf{500}&\textbf{400}& \textbf{250}	& \textbf{150}\\		
			\noalign{\smallskip}\hline
		\end{tabular}
\end{table} 

\section{Evaluation Metrics}

The task consists in classifying a news article as fake or real news. First, the training dataset was released for the participants to develop and train their systems, and subsequently, the test dataset was released. Each participant team had an option to submit only 3 different runs. The participants' submissions were evaluated by comparing the labels predicted by the participants' classifiers and the ground truth labels. To quantify the classification performance, we employed the commonly used evaluation metrics: Precision (P), Recall (R), Accuracy, and two F1-scores, namely, F1\textsubscript{real} for the prediction of label “real” and F1\textsubscript{fake} for the prediction of label “fake” out of all news. Additionally, to accommodate the skew towards the real class, which dominates (it has more samples than the fake news class), we used the macro-averaged F1-macro which is the average of F1\textsubscript{real} and F1\textsubscript{fake}.

As this is a binary classification problem and the dataset contains two equally important classes, we measured both of them, evaluating the quality of predicting whether a  news article is real, i.e., treating the “real” label as a positive, or target, class, and the quality of predicting whether the news article is fake, i.e.,  considering the label “fake” as a positive class. It is important to mention that since the dataset is not balanced, this is why these metrics are used.  

The final ranking is based on the F1-macro score. It can be observed that F1-macro  and accuracy are correlated, as it is expected.

\section{Baselines}
We provided three baseline systems with the goal that their performance could serve as reference points for qualitative evaluation of the submissions’ placement in the ranking. First, we provided the Random Baseline as the most basic and trivial baseline, which is expected to be ranked at the bottom with a more massive gap from the participating systems. Second, we provided the most traditional baseline: bag of words (BoW) model. It uses words as features and then apply a machine learning classifier. In this baseline, we used binary weighting scheme (i.e., a feature is present or not) with Logistic Regression classifier. For the third baseline, we provided the results of character bi-gram with tf-idf weighting scheme using Logistic Regression classifier, which achieved surprisingly good results. Overall, we tried five weighting schemes (tf-idf, logent, norm, binary, relative frequency) \cite{11} along with various classifiers such as Logistic Regression, SVM, Adaboost, Decision Tree, Random Forest, and Naive Bayes, but we got the best results with Logistic Regression, which we are reporting. 

\section{Overview of the Submitted Approaches}

This section gives a brief overview of the systems submitted to this competition. 42 teams registered for participation, 9 teams submitted their runs. Registered participants were from 6 different countries (India, Pakistan, China, Egypt, Germany, and the UK). This wide range of the regions where the interested participants were located confirms the importance of this task. The team members came from various types of organizations: universities, research centers, and industry.

As the initial step of the experimental setup, the majority of  participating teams performed data cleaning and preprocessing such as stop words elimination. In particular, the system submitted by the team MUCS removed the stopwords, whereas the systems submitted by teams BERT 4EVER, Chanchal\_Suman, CNLP-NITS, SSNCSE\_NLP and NITP\_AI\_NLP decided to leave them.

Further, all the news articles were represented with different text representation techniques. The team MUCS used traditional bags of words representation. Similarly, the teams MUCS, NITP\_AI\_NLP, and SSNCSE\_NLP used the n-gram (words or characters) representation weighted with tf-idf. The teams BERT 4EVER, CNLP-NITS, and Chanchal\_Suman represented news article texts using word embeddings. In particular, only one team, SSNCSE\_NLP, represented texts using Word2Vec embeddings, while the team SSNCSE\_NLP employed FastText embeddings. Furthermore, the team Chanchal\_Suman used Urdu word embedding and only one team, BERT 4EVER, used the contextual representation using BERT \cite{19}, which is one of the most recent and advanced manners of text representation.  

To implement their classifiers, some participating teams used the traditional, i.e., non-neural machine learning algorithms, while some teams submissions were based on various neural network architectures. The team MUCS used two classical machine learning classifiers such as Multinomial Naive Bayes and Logistic Regression with default parameters, and the same team also used LSTM in the experimental setup. Similarly, the team SSNCSE\_NLP used machine learning classifiers such as Multi-Layer Perceptron (MLP), AdaBoost (AB), ExtraTrees (ET), Random Forest (RF), Support Vector Machine (SVM), and Gradient Boosting (GB). Another team, NITP\_AI\_NLP, used ensemble models by combining  Random Forest, Decision Tree and AdaBoost classifiers. The  team  NITP\_AI\_NLP  also used a multi-layer dense neural network. In contrast, one team,  Chanchal\_Suman, used Gated Recurrent Unit (GRU). All the participating teams, except one team (NITP\_AI\_NLP), used  Transformers. Description of the approaches is presented in Table {\ref{dic}}. 

\begin{table}[!htb]
  \caption{Approaches used by the participating systems.}
  \resizebox{\columnwidth}{!}
  {%
   \label{dic}
  \begin{tabular}{ccccl}
    \toprule
System/Team Name & Feature Type & Feature Weighting Scheme & Classifying algorithm & NN-based\\
    \midrule
    
BERT 4EVER &   context embedding BERT &  BERT   &  CharCNN-Roberta   & Yes\\
Character bi-gram (baseline) &  char  bi-grams & TF-IDF   &  Logistic Regression  & No\\
BoW (baseline)&  word uni-grams & Binary   &  Logistic Regression  & No\\
CNLP-NITS &   $N/A$   &  embedding   &  XLNet pre-trained model     & Yes\\
NITP\_AI\_NLP &  char 1-3 grams   &  TF-IDF    &  Dense Neural Network   & Yes \\
Chanchal\_Suman &  $N/A$    & embeddings    & Bi-directional GRU model      & Yes \\
MUCS & mix of char and word n-grams    & embeddings    & ULMFiT model    & Yes\\
SSNCSE\_NLP & char n-gram & TFIDF, fastText, word2vec& RF, Adaboost, MLP, SVM    & Yes\\

  \bottomrule
\end{tabular}
}
\end{table}

\section{Results and Discussion}

Among all the submitted runs, the results of the best run (among up to three submitted runs) are presented in Table {\ref{tab:overlapping2}}. The systems are  ranked by the F1-macro score. Table 5 provides the aggregated statistics about the performance of all non-trivial systems, including the baselines, that is, all of the systems apart from the random baseline. 

We observe that except one system, all the other participating teams’ systems outperformed the random baseline in terms of F1-macro score. The BERT 4EVER system achieved the best F1-macro, Accuracy, as well as R\textsubscript{fake} (recall), F1\textsubscript{fake}, and P\textsubscript{real} (precision) scores. However, the baseline approach with character bi-grams and Logistic Regression achieved the second position in the shared task with just 1.1\% difference in F1-macro from BERT 4EVER, which is quite an unexpected result. The explanation of this fact is a question for further research. 

Table 4 presents the best results of the submitted systems.

\begin{figure}[h]
\includegraphics[width=\textwidth]{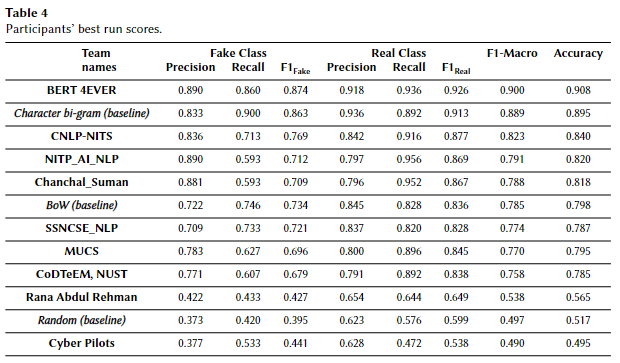}
\end{figure}

Over 50\% of the systems obtained F1-macro and Accuracy of more than 0.8 (Table  {\ref{tab:overlapping2}}), which is a reasonably high result.  In Table {\ref{dic}}, we observe that most of these high performing systems achieve better Recall for fake news, and better Precision for the real news detection. This tells us that these systems tend to “mistrust” the news articles and tag more news as fake than there are in reality. This can be considered as an overly secure approach.  

At this moment, it is hard to judge whether any of these approaches is ready to be applied “in the wild”. While the results of F1\textsubscript{real} and F1\textsubscript{fake} over 0.9 shown by the winning BERT 4EVER system are impressively high, the modest size of the provided training and testing datasets cannot guarantee the same performance on an arbitrary text input. To ensure the scalability of the presented approaches, more multifaceted research at a larger scale is needed. We see that one of the paths is a community-driven effort towards the increase of available resources and datasets in the Urdu language. 

Table {\ref{tab:freqs}} presents statistics of the submitted systems.

\begin{table}[!htb]
  \caption{Statistics  of the submitted systems and the baselines (excluding the random baseline).}
  \resizebox{\columnwidth}{!}
  {%
  \label{tab:freqs}
  \begin{tabular}{cccccccccl}
    \toprule
Stat. metrics & P\textsubscript{real} & R\textsubscript{real}& F1\textsubscript{real}& P\textsubscript{fake}&R\textsubscript{fake} & F1\textsubscript{fake}& F1-macro &Acc. \\
    \midrule
mean &  0.746 & 0.649 & 0.748 & 0.795 & 0.843 & 0.755 & 0.767 & 0.770\\
std&  0.193& 0.127& 0.131& 0.091& 0.161& 0.138& 0.130& 0.134\\
min&  0.377& 0.433& 0.490& 0.628& 0.472& 0.490& 0.502& 0.495&\\
percentile 10\%& 0.418& 0.523& 0.534&0.652& 0.627& 0.534& 0.560&0.558 \\
percentile 25\% &  0.725& 0.593& 0.762& 0.792& 0.838& 0.762& 0.781&0.786 \\
percentile 50\% &0.810&0.617&0.781&0.798&0.906&0.781&0.799&0.806 \\
percentile 75\% & 0.887&0.728&0.815&0.841&0.945&0.815&0.830&0.835\\
percentile 80\% & 0.890&0.745&0.828&0.850&0.949&0.839&0.846&0.850s\\
percentile 90\%& 0.891&0.800&0.850&0.888&0.952&0.902&0.882&0.892\\
max & 0.890&0.860&0.874&0.918&0.936&0.926&0.900&0.908
\\
  \bottomrule
\end{tabular}
}
\end{table}

\section{Conclusion}

This paper describes the first competition on automatic fake news detection in Urdu,  the UrduFake 2020 track at FIRE 2020. We provided the training and testing parts of the dataset that included news articles in five domains (business, health, sports, showbiz, and  technology). The news articles in the dataset were manually annotated with labels “fake” and  “real” with a slightly imbalanced ratio of approximately 60\% real news and 40\% fake news. 

Forty two teams from six different countries registered for this task. Nine teams submitted their experimental results (runs). The approaches employed by the submitted systems varied from the traditional feature-crafting and application of traditional ML algorithms to word representation through pre-trained embeddings to contextual representation and end-to-end neural network based methods. In particular, ensemble methods were used in the traditional ML case. LSTM, and Transformers (BERT) were used in neural network based solutions. 

Among all the submissions, only the best submitted model, BERT 4EVER, outperformed the character bi-grams with Logistic Regression baseline achieving F1-macro score of 0.90. This confirms that contextual representation and large neural network techniques outperform classical features-based models, which has been shown in many recent studies in all branches of natural language processing. 

 This competition aimed to encourage researchers working in different NLP domains  to attempt to tackle the proliferation of fake content on the web. It also provided an opportunity to fully explore the sufficiency of textual content modality and effectiveness of fusion methods. And last but not the least, this track provides a useful resource in the form of an annotated dataset for other researchers working in automatic fake news detection in Urdu. 

\begin{acknowledgments}

This competition was organized with partial support of National Council for Science and Technology (CONACYT) A1-S-47854, SIP-IPN 20200797 and 20200859 and CICLing conference. The work of the last author was partially funded by MICINN under the research project MISMIS-FAKEnHATE on MISinformation and MIScommunication in social media: FAKE news and HATE speech (PGC2018-096212-B-C31).
\end{acknowledgments}

%%
%% Define the bibliography file to be used
\bibliography{main}

\appendix

\end{document}